\documentclass[letterpaper]{article} 
\usepackage{aaai25}  
\usepackage{times}  
\usepackage{helvet}  
\usepackage{courier}  
\usepackage[hyphens]{url}  
\usepackage{graphicx} 
\usepackage{natbib}  
\usepackage{caption} 
\usepackage{algorithm}
\usepackage{algorithmic}
\usepackage{amsmath}
\usepackage{amsfonts}
\usepackage{booktabs} 
\usepackage{makecell} 
\usepackage{colortbl} 
\usepackage{array}
\usepackage{multirow}
\usepackage{centernot}
\usepackage{newfloat}
\usepackage{listings}
\DeclareCaptionStyle{ruled}{labelfont=normalfont,labelsep=colon,strut=off} 
\floatstyle{ruled}
\newfloat{listing}{tb}{lst}{}
\floatname{listing}{Listing}
\title{Order-Preserving Dimension Reduction for Multimodal Semantic Embedding}
\author{
    Chengyu Gong\equalcontrib$^1$,
    Gefei Shen\equalcontrib$^2$,
    Luanzheng Guo$^3$,
    Nathan Tallent$^3$,
    Dongfang Zhao$^4$
}
\affiliations{
    \textsuperscript{\rm 1}New York University, cg4761@nyu.edu\\
    \textsuperscript{\rm 2}Harvard University, gshen@g.harvard.edu\\
    \textsuperscript{\rm 3}Pacific Northwest National Laboratory, \{lenny.guo,nathan.tallent\}@pnnl.gov\\
    \textsuperscript{\rm 4}University of Washington, dzhao@uw.edu
}

\usepackage{bibentry}

\begin{document}

\maketitle

\begin{abstract}
Searching for the $k$-nearest neighbors (KNN) in multimodal data retrieval is computationally expensive, particularly due to the inherent difficulty in comparing similarity measures across different modalities. Recent advances in multimodal machine learning address this issue by mapping data into a shared embedding space; however, the high dimensionality of these embeddings (hundreds to thousands of dimensions) presents a challenge for time-sensitive vision applications. This work proposes Order-Preserving Dimension Reduction (OPDR), aiming to reduce the dimensionality of embeddings while preserving the ranking of KNN in the lower-dimensional space. One notable component of OPDR is a new measure function to quantify KNN quality as a global metric, based on which we derive a closed-form map between target dimensionality and key contextual parameters. We have integrated OPDR with multiple state-of-the-art dimension-reduction techniques, distance functions, and embedding models; experiments on a variety of multimodal datasets demonstrate that OPDR effectively retains recall high accuracy while significantly reducing computational costs.
\end{abstract}

%

\section{Introduction}
\label{sec:intro}

\subsection{Background and Motivation}

Multimodal retrieval tasks, especially those involving heterogeneous modalities such as text, image~\cite{ZHOU2015205}, and audio, rely on diverse data sources where conventional indexing techniques such as DBSCAN or $k$-means become ineffective. Many real-world multimodal systems, such as those in material science~\cite{RANGELDACOSTA2021103141}, often consist of multiple images and associated textual descriptions, stored in databases that manage structured (tables) and unstructured (blobs) data separately. Traditional indexing approaches fail to capture the unified semantic similarity between different modalities, requiring an alternative representation that aligns heterogeneous data within a common space.

Multimodal machine learning~\cite{li2022clipeventconnectingtextimages} provides a means to embed heterogeneous data into a shared vector space~\cite{jia2021scalingvisualvisionlanguagerepresentation}, enabling unified similarity search. However, embeddings from different modalities, such as text and images, are often concatenated, leading to an even higher-dimensional joint representation, which further exacerbates the curse of dimensionality~\cite{AREMU2020106706}. Given that embeddings generated by models such as BERT~\cite{devlin2018bert} and ViT~\cite{dosovitskiy2020image} already contain hundreds to thousands of dimensions, concatenation leads to prohibitively high-dimensional representations. This dimensionality explosion results in increased storage requirements, computational inefficiency, and degraded nearest-neighbor search performance, particularly in large-scale scientific applications~\cite{wu2020melindamultimodaldatasetbiomedical,tshitoyan2019unsupervised}.

Dimensionality reduction offers a potential solution by projecting high-dimensional embeddings into a lower-dimensional space while preserving key relationships. Techniques such as PCA~\cite{hotelling1933pca} and ISOMAP~\cite{tenenbaum2000isomap} aim to minimize the distortion in inter-point distances, facilitating efficient retrieval in reduced spaces. 
\textit{However, a fundamental challenge in applying dimensionality reduction to multimodal retrieval is the lack of a principled method to determine the optimal target dimensionality.} Without an a prior understanding of how well nearest-neighbor structures are preserved post-reduction, practitioners must rely on heuristic or empirical tuning, limiting the robustness of these methods in real-world applications~\cite{zhao2023retrievingmultimodalinformationaugmented}. This uncertainty is particularly problematic in tasks such as cross-modal retrieval~\cite{liang2024multi}, Retrieval-Augmented Generation (RAG)~\cite{lewis2020retrieval}, and scientific knowledge discovery~\cite{gupta2022matscibert,lu2023multiscale}, where the integrity of neighborhood relationships directly impacts retrieval effectiveness.

These challenges underscore the necessity of understanding the trade-off between dimensionality reduction and nearest-neighbor preservation. A framework that enables quantitative estimation of neighbor preservation after dimensionality reduction would allow for more effective multimodal retrieval by mitigating the curse of dimensionality while maintaining computational efficiency.

\subsection{Proposed Work}

To effectively reduce the dimensionality of high-dimensional embedding vectors while preserving the neighborhood structure, we introduce a mathematically rigorous measure function that quantifies the preservation of nearest neighbors across different metric spaces. This measure, formally defined over the power-set $\sigma$-algebra of the lower-dimensional space, quantifies the degree to which $k$-nearest neighbors are preserved. By enabling a well-defined aggregation over the entire space, this measure provides a concrete way to evaluate the fidelity of dimension-reduction transformations.

Building on this measure, we analyze the quantitative relationship between neighborhood preservation and target dimensionality. We construct a closed-form function that captures the intrinsic dependencies among key parameters, including the original and reduced dimensionality, the number of data points, and the expected neighborhood preservation accuracy. Our analysis reveals that the required lower-dimensional space is positively correlated with both the original dimensionality and dataset cardinality, but their influence varies. In particular, neighborhood preservation accuracy exerts an exponentially stronger effect than dataset size, indicating its dominant role in determining the optimal reduced dimensionality.

In addition to theoretical results, we demonstrate how the proposed measure and closed-form function integrate into practical vision applications. We detail their incorporation into dimensionality reduction pipelines using PCA~\cite{hotelling1933pca} and MDS~\cite{kruskal1978mds}, evaluate their compatibility with state-of-the-art embedding models such as CLIP~\cite{radford2021learningtransferablevisualmodels}, ViT~\cite{dosovitskiy2020image}, and BERT~\cite{devlin2018bert}, and examine their performance under different distance metrics, including Euclidean, cosine, and Manhattan. Extensive experiments on multimodal scientific datasets and image-text benchmarks validate the accuracy of the closed-form function in capturing the relationship between dimensionality reduction and neighborhood preservation. The results highlight the effectiveness of this framework in mitigating the curse of dimensionality while ensuring robust retrieval performance in vision-based applications.

\paragraph{Contributions}
This work makes the following key contributions:
\begin{itemize}
    \item We introduce a mathematically rigorous measure function to quantify the preservation of $k$-nearest neighbors across different dimensional spaces, providing a formal foundation for evaluating dimensionality reduction methods.
    
    \item We derive a closed-form function that quantitatively describes the relationship between neighborhood preservation and target dimensionality, offering a principled approach to determining the optimal reduced dimensionality.
    
    \item We empirically validate the proposed framework across multiple datasets and dimensionality reduction techniques, demonstrating that the closed-form function accurately predicts neighborhood preservation and effectively mitigates the curse of dimensionality.
\end{itemize}

\section{Background and Related Work}


\subsection{Multimodal Data Analytics}
Multimodal data analytics  has attracted growing research interest across scientific domains. For example,  the MELINDA dataset \cite{wu2020melindamultimodaldatasetbiomedical} focuses on classifying biomedical experiment methods by utilizing multiple data types. Similarly, a study on explainable AI \cite{Jin_2022} evaluates algorithms within the context of multimodal medical imaging tasks, emphasizing the importance of decision-making based on raw scientific data. In parallel, researchers have developed new large language models specifically designed for scientific data. For example, GIT-Mol \cite{LIU2024108073} represents a notable effort, developing a multimodal large language model for molecular science that integrates graph, image, and text data. Additionally, MatSciBERT \cite{gupta2022matscibert} focuses on creating domain-specific language models tailored to materials science. While these models, along with others such as the unsupervised word embeddings for materials science literature \cite{tshitoyan2019unsupervised}, demonstrate strong potential in capturing domain-specific knowledge, they are often limited in scope, typically focusing solely on text-based data.

Multimodal data is often highly dimensional, making it crucial to find a trade-off between maintaining essential information and reducing the data's complexity. 
Given the high dimensionality of multimodal data, researchers have proposed methods to balance information preservation and complexity reduction.
For instance, the study on Multiscale Feature Extraction and Fusion of Image and Text in VQA \cite{lu2023multiscale} introduces advanced techniques to fuse multimodal data, allowing for a more integrated analysis of different data types. 
The research on Semi-Supervised Multimodal Learning with Balanced Spectral Decomposition \cite{hu2020semi} focuses on optimizing information preservation through spectral decomposition, which enhances the ability to capture correlations within the data. 
Additionally, Unsupervised word embeddings in materials science \cite{tshitoyan2019unsupervised} demonstrates the transformation of textual information into vectors, providing a compact and efficient representation of large-scale text data.

\subsection{Contrastive Language-Image Pre-Training}

With the introduction of ``Attention is All You Need''~\cite{vaswani2017attention}, the development of large language models has surged, leading to significant advancements in the field. Following this breakthrough, models like BERT \cite{devlin2018bert} and ViT \cite{dosovitskiy2020image} have focused on text analysis and image recognition respectively. Particularly noteworthy is the CLIP model \cite{radford2021learningtransferablevisualmodels}, which integrates both text and image data, marking a significant step forward in multimodal data processing. Various improvements to the CLIP model have been proposed. For instance, TiMix \cite{jiang2024timixtextawareimagemixing}, address issues with noisy web-harvested text-image pairs by using mixed data samples, while CLIP-Event \cite{li2022clipeventconnectingtextimages} and SoftCLIP \cite{Gao_Liu_Xu_Wu_Zhang_Li_Yang_Liu_Sun_2024} further enhance the alignment between text and images.

In~\cite{zhang2021probability}, authors proposed to convert images into binary codes (hash codes) that preserve image similarity, making it easier to compare images within large-scale datasets.

Some prior works proposed methods to perform dimension reduction over the embedding vectors. For example, in~\cite{NEURIPS2022_297f7c6c}, authors proposed a new approach to reducing dimensionality following contrastive learning.  Similarly, in~\cite{huang2020improve}, authors focused on relative positioning in data before and after LLM manipulation.
However, none of the above works touched on the preservation of the  set of $k$-nearest neighbors. 

\subsection{Dimension Reduction}
After Multidimensional Scaling (MDS) \cite{torgerson1952multidimensional} was published, numerous methods were proposed that preserved pairwise distances in lower-dimensional spaces. 
For example, Colored Maximum Variance Unfolding \cite{NIPS2007_55a7cf9c} preserved local distances while maximizing variance, and Neighborhood Preserving Embedding \cite{he2005neighborhood} retained local relative positions.
Further innovations include Tensor Embedding Methods \cite{dai2006tensor}, which tackled the curse of dimensionality, and MultiMAP \cite{jain2023multimap}, which integrated multimodal data. Recent approaches like Similarity Order Preserving Discriminant Analysis \cite{hu2021novel} and Ordinal Data Clustering Algorithm with Automated Distance Learning \cite{zhang2020ordinal} specifically aimed to preserve data order post-reduction, highlighting the evolving focus on maintaining data integrity across various dimensions. 
However, none of the above works considered preserving the \textit{set} of $k$-nearest neighbors during the dimension reduction.

As large language models (LLMs) become popular, scientists have also proposed methods that integrate LLMs with dimension-reduction techniques to enhance the analysis of complex data. For example, Transformer-based Dimensionality Reduction \cite{ran2022transformerbaseddimensionalityreduction} introduced a method that decomposes autoencoders into modular components, leveraging the power of transformers to manage high-dimensional data efficiently. Similarly, in~\cite{george2023integrated}, authors combined BERT with dimensionality reduction to enhance topic modeling by clustering textual data more effectively.  
However, the above works assumed that the users of LLMs were able to modify the model structure,
which is unrealistic in many scientific applications.

\subsection{Multimodal Embedding and Retrieval across Heterogeneous Data}
Recent progress in multimodal representation learning has significantly advanced cross-modal retrieval tasks, which aim to identify semantically aligned items across heterogeneous modalities such as text, images, and audio.
Traditional vector retrieval methods, including FAISS\cite{douze2024faiss}, ScaNN\cite{guo2020accelerating}, and HNSW~\cite{malkov2018efficientrobustapproximatenearest}, enable efficient similarity search for high-dimensional embeddings. However, as embedding dimensionality increases, these methods face scalability challenges, leading to degraded retrieval efficiency and increased computational cost. While existing approaches focus on scaling similarity search or enhancing retrieval precision, few address the structural integrity of nearest-neighbor relations after dimensionality reduction. To address this limitation, our work introduces Order-Preserving Dimension Reduction (OPDR), which reduces embedding dimensionality while preserving $k$-nearest neighbor (kNN) structure, thereby improving retrieval efficiency in large-scale multimodal datasets.

\section{Order-Preserving Dimension Reduction}

This section first presents a new notion,
i.e., Order-Preserving Measure (OPM), 
for the \textit{set} of $k$-nearest neighbors that are preserved during a map between two metric spaces.
Then, a closed-form function is constructed to quantify the relationship among the space dimensionality, the number of data points, and the OPM.
Finally, a detailed implementation is presented about how to incorporate the new measure and function real-world multimodal retrieval systems.

\subsection{Order-Preserving Measure}
\label{sec:opdr_measure}

Informally, an order-preserving measure (OPM) provides a metric of the number of nearest neighbors that do not change between two spaces.
For example, if the $2$-closest points (we assume that the spaces are metric spaces such that pair-wise distances are well defined) of $every$ point in a metric space $(X, \delta_X)$ are still the same $2$-closest points (of each point) in a metric space $(Y, \delta_Y)$,
then we say the map $f: X \to Y$ is order-preserving of 2, or $OP_2$.
We will provide a more formal definition of $OP_z$, $z \in \mathbb{Z}^+ \cup \{0\}$ later;
before that, we need to point out a common misunderstanding of this notion regarding inclusiveness, as follows.

The order-preserving notion defined above is \textit{not} inclusive in the sense that in general,
$OP_{k+1} \centernot\implies OP_k$, $1 \le k \in \mathbb{Z}$.
This can be understood as the \textit{set} of top $k$ nearest neighbors in the space $Y$ do not necessarily respect the intrinsic order of elements in the \textit{set} of the top $k$ nearest neighbors in space $X$.
The rationale behind this is that the result of a $k$-nearest neighbor (KNN) query in many cases will be the input of other analytical steps that are agnostic of the ``internal'' order of the set of points.
This implies that, for example, an $OP_2$ map is not necessarily $OP_1$:
A sorted list of points in space $Y$, $L_Y = (b, a, c)$ is clearly $OP_2$ if the original sorted list on space $X$ is $L_X = (a, b, c)$ because $\{b, a\} = \{a, b\}$ (even if $(b, a) \not= (a, b)$ as ordered lists); 
However, $L_Y$ is \textit{not} $OP_1$ regarding $L_X$ because $\{b\} \not= \{a\}$.
For completeness, we will agree that $OP_0$ is trivially true for any pairs of lists.

Formally, a \textit{measure} is a function, say $\mu$, which maps a subset $E_i$ of set $X$ in a well-defined $\sigma$-algebra to a value in the extended real line $[0, \infty]$, such that (i) $\mu(\emptyset)$ = 0 and (ii) $\mu\left(\bigcup_{i=1}^n E_i\right) = \sum_{i+1}^n \mu(E_i)$.
Because we assume there exists a well-defined $\sigma$-algebra over (the set of) embedding vectors, say $X$,
$E_i$'s will be understood as \textit{disjoint} subsets of $X$ unless otherwise stated.
In fact, we will explicitly construct the $\sigma$-algebra on the target space, $Y$.
As a side note, a measure function $\mu$ can be also thought of as a \textit{homomorphism} between the two monoids (i.e., groups without inverses) $(\mathcal{P}(X), \cup)$ and $(\mathbb{R}^+ \cup \{0\}, +)$ with kernel $Ker_\mu = \{\emptyset\} \subseteq \mathcal{P}(X)$,
where $\mathcal{P}(X)$ denotes the power set of $X$.
However, our analysis in the following will not require any group-theoretical results.

The construction of the $\sigma$-algebra of embedding vectors is as follows.
Let $k$ denote the target number of preserved $k$-nearest neighbors.
Let $Y$ denote the target (i.e., low-dimensional) metric space (we omit the metric functions here) of the dimension-reduction map.
In practice, the cardinality of $Y$ is finite (i.e., there is an upper limit of number of embedding vectors in vision applications), 
which is of course countable.
The $\sigma$-algebra $\mathcal{M}_Y$ of $Y$ is simply the power set of all the mapped vectors in $Y$,
i.e., $\mathcal{M}_Y = \mathcal{P}(Y)$,
which is obviously a $\sigma$-algebra on $Y$ because $Y$ is countable and $\mathcal{P}(Y)$ is closed for any countable number of set unions.

After constructing its $\sigma$-algebra,
we are ready to define the measure on $\mathcal{M}_Y$.
Let 

$E^Y_{k,i}$ denote the set of the $k$-nearest neighbors of $y_i$, where $y_i \in Y$.
The same notation can be defined for space $X$,
which denotes the original metric space where the original vectors of $y_i$ reside,

i.e., $x_i = f^{-1} (y_i)$, where $f$ denotes the dimension-reduction function.
Let $\mu$ be a function defined as follows,
\begin{equation}\label{eq:measure}
\displaystyle
\mu(F_j) = \frac{\left|F_j \cap E^Y_{k,i} \cap E^X_{k,i}\right|}{k},
\end{equation}
where $F_j \in \mathcal{M}_Y$.
It should be clear that $\mu \in [0, 1] \subset [0, +\infty]$ because $\left| E^Y_{k,i}\right| \in [0, k]$, where $|\cdot|$ denotes the cardinality of a set. 

We argue that the function defined in Eq.~\eqref{eq:measure} is a measure on $\mathcal{M}_Y$.
This means that we need to demonstrate two properties of $\mu$ (on $\mathcal{M}_Y$):
(i) $\mu(\emptyset) = 0$ and (ii) $\mu(\bigcup_{i=1}^\infty F_j) = \sum_{i=1}^\infty (\mu(F_j))$, $\forall F_j \subseteq Y$ and $F_j$'s are disjoint.
Property (i) is trivially satisfied because when $F_j = \emptyset$,
we have $\mu(F_j) = \frac{\left|\emptyset\right|}{k} = \frac{0}{k} = 0$.
To prove property (ii), it suffices to show that if $F_j = F_1 \cup F_2$ and $F_1 \cap F_2 = \emptyset$, then $\mu(F_1 \cup F_2) = \mu(F_1) + \mu(F_2)$;
this is because $\mathcal{M}_Y$ is finite and the above binary relationship can be extended finite times.
To simplify the notation, let $E = E^Y_{k,i} \cap E^X_{k,i}$.
It follows that $\mu(F_j)$ = $\frac{\left|(F_1 \cup F_2) \cap E\right|}{k}$ = $\frac{\left|(F_1 \cap E) \cup (F_2 \cap E)\right|}{k}$.
Now, because $F_1 \cap F_2 = \emptyset$, we know $(F_1 \cap E) \cap (F_2 \cap E) = \emptyset$.
This implies that there is no common element between $(F_1 \cap E)$ and $(F_2 \cap E)$, which means the sets are separate components and the cardinality of the union set is simply the addition of cardinality of each component:
$\left|(F_1 \cap E) \cup (F_2 \cap E)\right| = \left|F_1 \cap E\right| + \left|F_2 \cap E\right|$.
It follows that $\mu(F_1 \cup F_2) = \frac{\left|F_1 \cap E\right| + \left|F_2 \cap E\right|}{k} = \frac{\left|F_1 \cap E\right|}{k} + \frac{\left|F_2 \cap E\right|}{k} = \mu(F_1) + \mu(F_2)$, as desired.

The discussion about preserving the set of $k$-nearest neighbors between metric spaces around Eq.~\eqref{eq:measure} is, in our humble opinion, a tip of the iceberg regarding the \textit{neighborhood-preserving dimension-reduction maps}.
The reasoning is that the $k$-nearest neighbors can be thought of the discrete representation of the neighborhood of the referred point, or vector.
If we extend the idea to a continuous counterpart,
i.e., given a point $p$ and its neighborhood or an \textit{open set} around it, say $\epsilon(p)$,
the question becomes under what conditions the open set of the mapped vectors, 

namely $f(p)$, corresponds to the open set of $p$ in a dimension-reduction function $f$.
In topology and analysis, such a function carrying the open sets in the forward direction is called an \textit{open map},
which is unfortunately not the same (in fact, the inverse) condition under which the function is continuous. 
We leave this as an open question to the community.

\subsection{Closed-Form Function}
\label{sec:opdr_function}

The previous section defines an additive measure $\mu$ on the $\sigma$-algebra of the projected space $Y$;
This section investigates the relationship between $\mu$ and other parameters, such as the space dimensionality and the number of data points, or space cardinality.
The idea is to construct a function for the above variables which we hypothesize to be critical for dimensionality reduction, largely based on our empirical observations and mathematical intuition.
We will later verify the constructed function as a working hypothesis in the evaluation section.

We assume that we have an existing dimension-reduction method on hand and we are interested in quantifying the target dimensionality of the lower-dimensional space in which the set of $k$-nearest neighbors is an \textit{invariant}, as defined in Eq.~\eqref{eq:measure}.
It turns out that before attempting to solve the problem,
we need to have a more ``global'' metric than $\mu(\cdot)$, which works as a local measure that is concerned with only the $k$-nearest neighbors of a single point.

We define the \textit{global} metric for quantifying the overall closeness (or, similarity, accuracy) between two metric spaces regarding their $k$-nearest neighbors as follows.
We first aggregate the measure for each point,
then normalize the aggregate measure (AM) into $[0, 1]$,
and finally calculate the arithmetic mean of all normalized AMs (NAMs).
We call the above average of NAMs as the \textit{accuracy} of $Y$ with respect to $X$ on $k$-nearest neighbors.
Formally, we define the accuracy $A$ as
\begin{equation}\label{eq:acc}
\displaystyle
\text{A}^X_k(Y) = \frac{1}{m} \cdot \sum_{i=1}^m \frac{\mu_i\left(Y \setminus \{y_i\}\right)}{k},
\end{equation}
where $m = \left|Y\right| = \left|X\right|$, $y_j \in Y$, and $\mu_i(\cdot)$ is the measure on $Y$ over $y_i$ as defined in Eq.~\eqref{eq:measure}.
It is not hard to see that the accuracy $A$ defined above falls within the range $[0, 1]$,
because $\mu_i\left(Y \setminus \{y_i\}\right)$ is bounded by $k$ (inclusively).

The parameters in our hypothetical function include the accuracy $A_k$ between two metric spaces $X$ and $Y$, the number of data points $m = \left|X\right| = \left|Y\right|$,
and the dimensionality of the spaces.
In practice, the dimension of the domain space $X$ is determined by the neural network model;
therefore, our function would assume that $X$'s dimension is a constant and only involves the dimension of $Y$, denoted by $dim(Y)$.
Thus, we expect to construct a function $g$ as follows,
\[
dim(Y) = g\left( A_k, m \right),
\]
such that the result $dim(Y)$ can be set as a parameter in a dimension-reduction function $f$.
It follows that the real-world users can simply compose the functions $g$ and $f$,
i.e., $f\circ g$,
to ensure that the set of $k$-nearest neighbors is an invariant between two spaces $X$ and $Y$.

The construction of $g$ is based on the following observations:
\begin{itemize}
    \item Firstly, $dim(Y)$ is positively influenced by both the accuracy $A_k$ and the cardinality $m$.
    That is, if the user expects to maintain a higher accuracy of the $k$-nearest neighbors in a lower-dimensional space $Y$, then the target dimension $dim(Y)$ should be also set higher. In the extreme case, if $Y = X$, then $A_k = 1.0$.
    Similarly, if there are a large quantity of data points, 
    then the intuition is that we may need to keep a large number of dimensions in the target low-dimensional space $Y$.

    \item Secondly, the impact of the accuracy $A_k$ should be higher than that of the cardinality $m$.
    This can be understood with the intuition that $dim(Y)$ is quite sensitive to $A_k$ because $A_k$ is a global metric of the entire space $Y$.
    On the other hand, a small change of the number of data points may or may not significantly change the distribution of the space.
\end{itemize}
Based on the above discussion, 
we postulate that the function $g$ takes the following form
\begin{equation}\label{eq:dim}
\displaystyle
    g(a, b) = \mathcal{O}(b\cdot 2^a),
\end{equation}
where the exponential factor $2^a$ underscores the significantly higher impact of $a$.
In other words, the closed-form function is in the form of
\[
dim(Y) = \mathcal{O}(m\cdot 2^{A_k}).
\]
If we simplify the asymptotic notation with constant coefficients,
we can also rewrite the function as
\begin{equation}\label{eq:gfunc}
A_k = c_0\cdot \log \frac{dim(Y)}{m} + c_1,
\end{equation}
where $c_0$ and $c_1$ are two constant values that can be estimated by various regression models.
When $A_k = 1$, we say the low-dimensional space $Y$ is $OP_k$ to the original space $X$, 
which formalizes our previous notion of $OP_K$.

In the following sections, we will verify the effectiveness of the function defined in Eq.~\eqref{eq:gfunc};
but before that, we will describe how we implement the measure $\mu(\cdot)$ and the dimension-reduction function $f$ guided by Eq.~\eqref{eq:gfunc}.

\subsection{Integration into Multimodal Data Retrieval}
We will first describe how the embedding vectors are generated from multimodal  data, specifically the image-text pair and then present how we implement the hypothetical function and incorporate it into popular dimension-reduction methods.

To leverage transformer-based models, including CLIP (Contrastive Language-Image Pretraining), Vision Transformer (ViT), and BERT (Bidirectional Encoder Representations from Transformers), for converting multimodal data into embedding vectors, we conducted experiments on three distinct datasets: materials science data, Flickr30k, and OmniCorpus. Each dataset consists of multimodal data, incorporating text from HDF5 files, natural language text, and images in TIFF, PNG, or JPEG formats. The experimental results demonstrate that our approach maintains strong applicability across different domains, including both scientific data and natural images.

The generated embeddings retain the default dimensionality of each respective model: BERT and ViT both produce 768-dimensional embeddings. For CLIP, the text and image encoders each output 512-dimensional vectors. We construct unified multimodal representations by directly concatenating embeddings from different modalities—for example, combining CLIP text and image vectors into a single 1024-dimensional vector.

In the case of audio-text data from the ESC-50 dataset, we generate embeddings using BERT (768D) for textual labels and PANNs CNN14~\cite{kong2020pannslargescalepretrainedaudio} (2048D) for audio signals. These are concatenated into 2816-dimensional joint vectors.
All embeddings—spanning image, text, and audio modalities—are stored for subsequent dimensionality reduction and retrieval analysis.

Following the embedding vector extraction, we focus on the dimensionality reduction phase. We evaluated several mainstream techniques including PCA, and MDS, among which PCA consistently outperformed others in terms of maintaining the integrity of location information in our datasets. Our investigations revealed a notable correlation between the effectiveness of PCA in preserving the spatial relationships and the ratio 

$\frac{n}{m}$, where $n = dim(Y)$ and $m = \left|Y\right|$,
of the target dimension to the number of samples used in the PCA process. We adopted various regression models to elucidate the relationship between the accuracy of preserved relative location information and the ratio $\frac{n}{m}$. These models facilitate the prediction of an optimal embedding vector dimension required to achieve a predetermined accuracy level, given a known number of samples, $m$.

\section{Evaluation}

This section evaluates the effectiveness of the proposed method for KNN-preserved dimension reduction on multiple data sets.
To ensure the robustness of our method across various scenarios, we evaluate the method with several Transformer-based models for extracting embedding vectors, 
two popular dimension-reduction techniques, 
and three distance metrics.
All results suggest that the proposed method is highly effective.

\subsection{Experimental Setup}

\paragraph{Data Sets}
The evaluation was conducted using seven datasets: four scientific datasets from the Materials Project~\cite{material_project}, two natural image-text pair datasets (Flickr30k~\cite{young2014image} and OmniCorpus-037 CC~\cite{li2024omnicorpus}), and one audio-text dataset (ESC-50~\cite{10.1145/2733373.2806390}).

Specifically, the observable, stable, metal, and magnetic datasets contain 33,990; 48,884; 72,252; and 81,723 data points, respectively. Flickr30k includes 31,014 image-text pairs, while OmniCorpus-037 CC provides 3,878,063 such pairs. The ESC-50 dataset comprises 2,000 environmental audio clips, each paired with a textual label describing the sound event (e.g., “dog barking”, “siren”, “rain”). 
\paragraph{Platform}
Our experiments were carried out on the NSF-sponsored computing platform CloudLab~\cite{cloudlab}, specifically the \textit{d7525} nodes with the following specifications. The machine is equipped with two 16-core AMD EPYC 7302 CPUs at 3.00GHz, 128GB ECC Memory (8×16 GB 3200MT/s RDIMMs), and two 480 GB 6G SATA SSDs along with one 1.6 TB PCIe4 x4 NVMe SSD for storage. The network interface card (NIC) is a dual-port Mellanox ConnectX-6 DX 100Gb NIC, with one port supporting 200Gb connectivity for experimental use. Additionally, the machine is equipped with an NVIDIA 24GB Ampere A30 GPU.

We have installed the following libraries in conda environment: transformer, sklearn, torch, pandas, numpy, os, pickle, h5py, shutil, matplotlib, and pyprismatic, PIL.

\subsection{OPDR on Various Data Sets}
\begin{figure}[!t]
    \centering
    \begin{minipage}{0.23\textwidth}
        \centering
        \includegraphics[width=\textwidth]{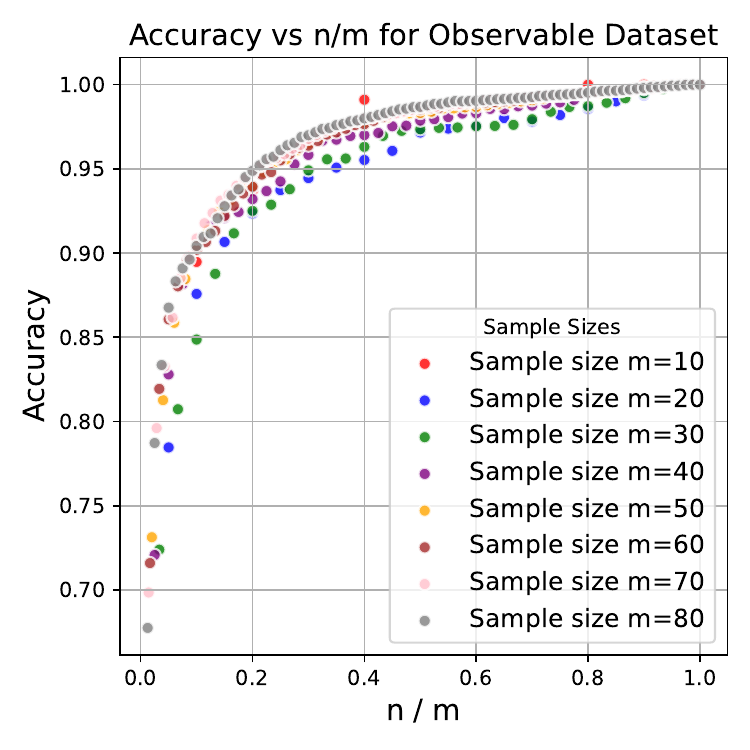}
        \caption{\small Observable 
        }
        \label{fig:observed_M}
    \end{minipage}\hfill
    \begin{minipage}{0.23\textwidth}
        \centering
        \includegraphics[width=\textwidth]{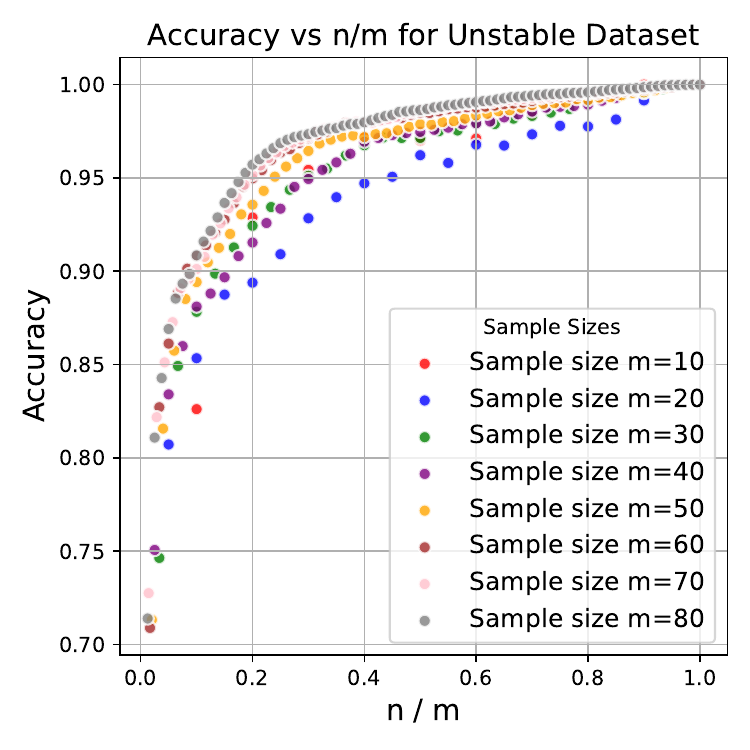}
        \caption{\small Unstable}
        \label{fig:unstable_M}
    \end{minipage}\hfill
    \begin{minipage}{0.23\textwidth}
        \centering
        \includegraphics[width=\textwidth]{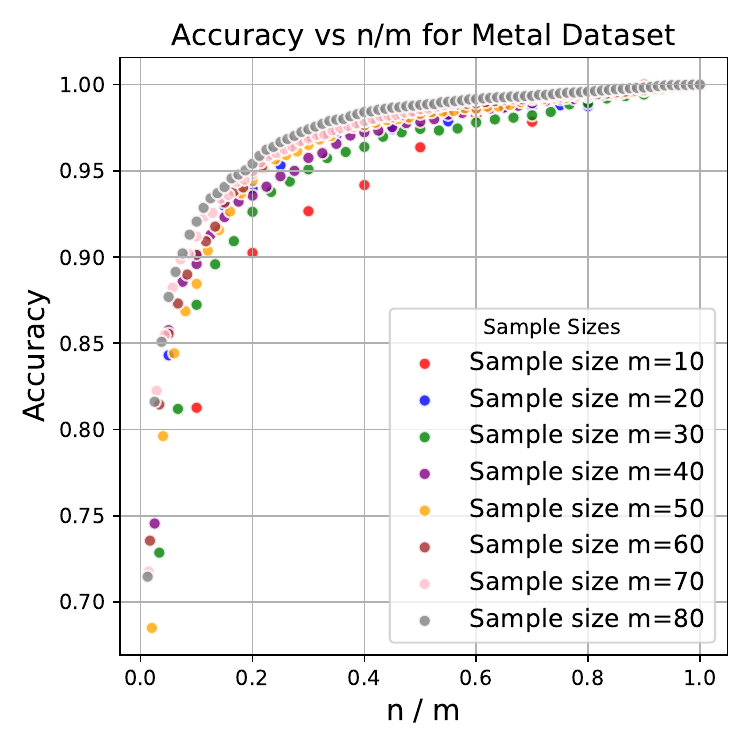}
        \caption{\small Metal}
        \label{fig:metal_M}
    \end{minipage}\hfill
    \begin{minipage}{0.23\textwidth}
        \centering
        \includegraphics[width=\textwidth]{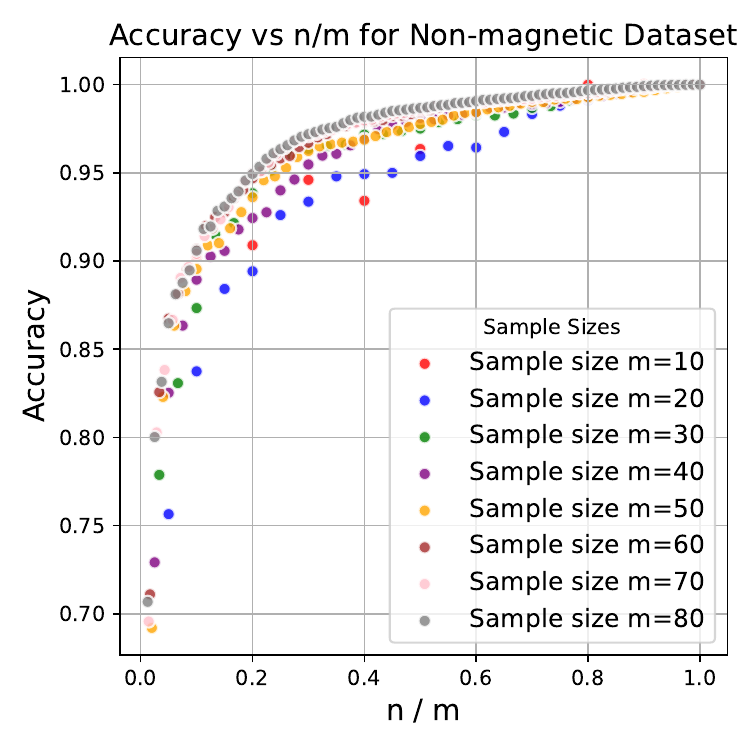}
        \caption{\small Non-magnetic}
        \label{fig:non-magnetic_M}
    \end{minipage}
    \begin{minipage}{0.23\textwidth}
        \centering
        \includegraphics[width=\textwidth]{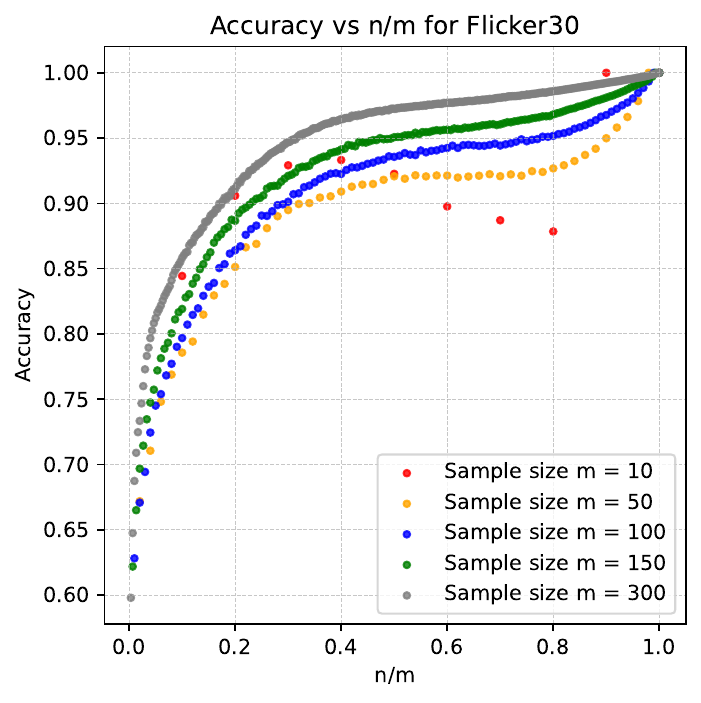}
        \caption{\small Flicker30k}
        \label{fig:Flicker30_pca_clip}
    \end{minipage}
      \begin{minipage}{0.23\textwidth}
        \centering
        \includegraphics[width=\textwidth]{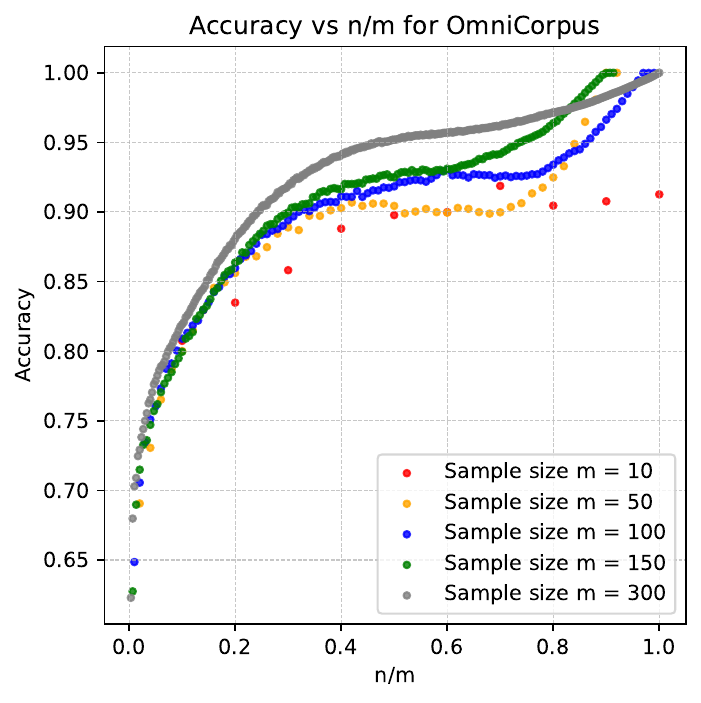}
        \caption{\small OmniCorpus}
        \label{fig:OmniCorpus_pca_clip}
    \end{minipage}
\end{figure}

We first use the CLIP model to extract embedding vectors from six datasets, comprising four scientifically curated multimodal datasets (Observable, Unstable, Metal, Non-magnetic), two widely adopted vision-language benchmarks (Flickr30k and OmniCorpus), as well as the ESC-50 dataset~\cite{10.1145/2733373.2806390} for audio-text representation. Subsequently, we computed the distances between vectors using the L2-norm and performed dimensionality reduction using PCA. To simplify notation, we let $n = dim(Y)$, denote the dimension of the lower-dimensional space. For these four groups of material datasets, we further divided them into eight 
subsets each, with sample sizes \( m \) where \( m \in \{10, 20, 30, 40, 50, 60, 70, 80\} \). For the remaining two multimodal datasets, we divided them into five subsets each, with sample size  \( m \) where \( m \in \{10, 50, 100, 150, 300\} \), to test whether the distribution patterns of data points remain consistent across different sample sizes.
Observations from Figure~\ref{fig:observed_M} to Figure~\ref{fig:OmniCorpus_pca_clip} indicate a strong positive correlation between accuracy and the ratio of \( n \) to \( m \). As \( n \) approaches \( m \), accuracy initially increases rapidly and then slows down, converging to a stable value. Although the four different datasets exhibit highly similar data distributions, minor differences may be attributed to the inherent distribution characteristics and randomness of the samples.

Through comparative analysis of the six data sets discussed above, we observe that our proposed OPDR algorithm performs consistently across multiple data sets and exhibits the anticipated data patterns, validating our initial hypotheses. 
In the remainder of this section, we will focus on the Observable Material dataset, exploring the effects of using different Transformer-based models, dimensionality reduction techniques, and various distance metrics.

\subsection{Influence of Embedding Models}

To evaluate the impact of different embedding models, we conducted further experiments on the Observable dataset, alongside Flickr30k and OmniCorpus from the multimodal datasets, using different models for extracting embedding vectors. We employed three distinct transformer-based models—BERT (Bidirectional Encoder Representations from Transformers), Vision Transformer (ViT), and CLIP (Contrastive Language-Image Pre-training)—to process and analyze the datasets. This enabled us to obtain more holistic and unified representations across different data domains.  

Through extensive experiments with these three models, we derived data fit lines to represent the relationship between accuracy and the ratio \( n/m \). All three fit lines consistently indicated that as \( n \) approaches \( m \), accuracy initially increases rapidly and then gradually converges. The results from the three different models aligned with our initial hypotheses.  Additionally, for datasets with structured relationships—such as material-related datasets—the obtained fit lines were nearly overlapping, suggesting that the choice of model does not significantly alter the underlying data structure. In contrast, for the Flickr30k and OmniCorpus datasets, differences between models were more pronounced, though the overall trend still followed the expected logarithmic pattern. This indicates that while different transformer-based models may introduce variations in representation, they maintain the fundamental structure of the data.  These findings reinforce that, regardless of the chosen model, the core data relationships remain stable. Thus, in practical applications, selecting the most appropriate model should be guided by the characteristics of the dataset and the specific task requirements, rather than concerns over fundamental data integrity.

\begin{figure}[!t]
    \centering
    \begin{minipage}{0.23\textwidth}
        \centering
        \includegraphics[width=\textwidth]{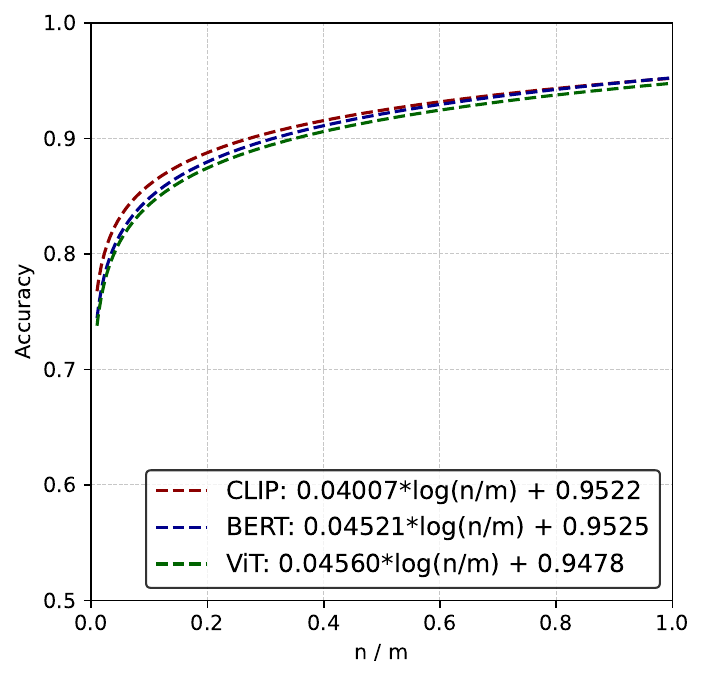}
        \caption{\small Data Fitting for Embedding Models on Material Dataset
        }
        \label{fig:llm_fit}
    \end{minipage}\hfill
    \begin{minipage}{0.23\textwidth}
        \centering
        \includegraphics[width=\textwidth]{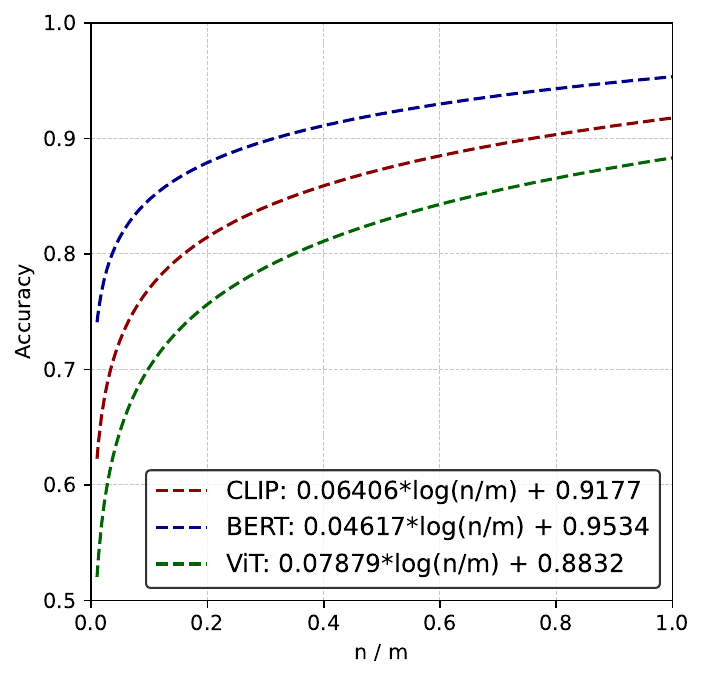}
        \caption{\small Data Fitting for Embedding Models on Flicker dataset}
        \label{fig:Flicker_model}
    \end{minipage}\hfill
    \begin{minipage}{0.23\textwidth}
        \centering
        \includegraphics[width=\textwidth]{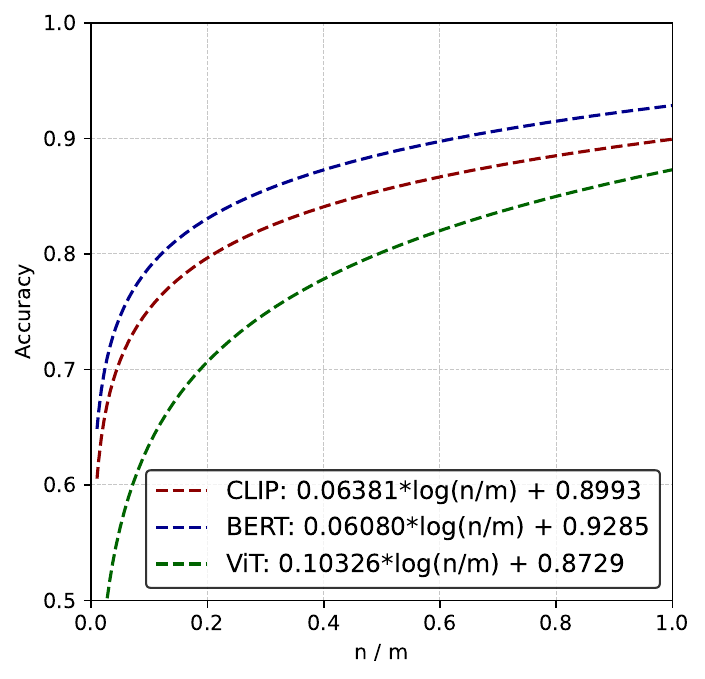}
        \caption{\small Data Fitting for Embedding Models on OmniCorpus dataset}
        \label{fig:OmniCorpus_model}
    \end{minipage}\hfill
    \begin{minipage}{0.23\textwidth}
        \centering
        \includegraphics[width=\textwidth]{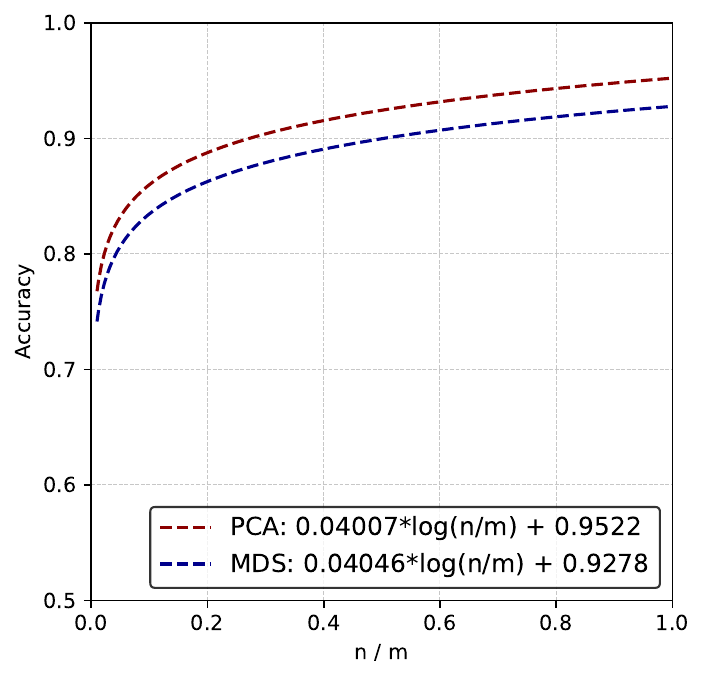}
        \caption{\small Data Fitting for Dimension Reduction on Material Dataset}
        \label{fig:DR}
    \end{minipage}
    \hfill
    \begin{minipage}{0.23\textwidth}
        \centering
        \includegraphics[width=\textwidth]{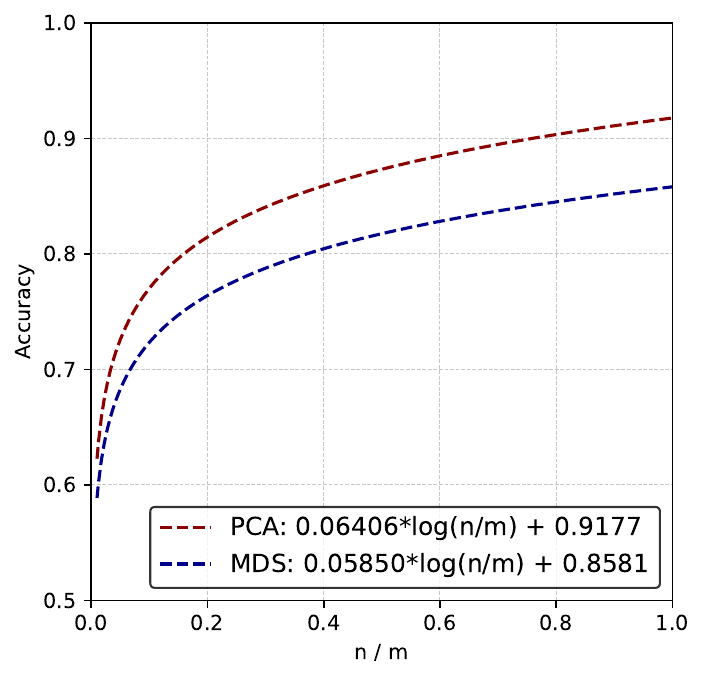}
        \caption{\small Data Fitting for Dimension Reduction on flicker dataset}
        \label{fig:DR_flicker}
    \end{minipage}
    \hfill
      \begin{minipage}{0.23\textwidth}
        \centering
        \includegraphics[width=\textwidth]{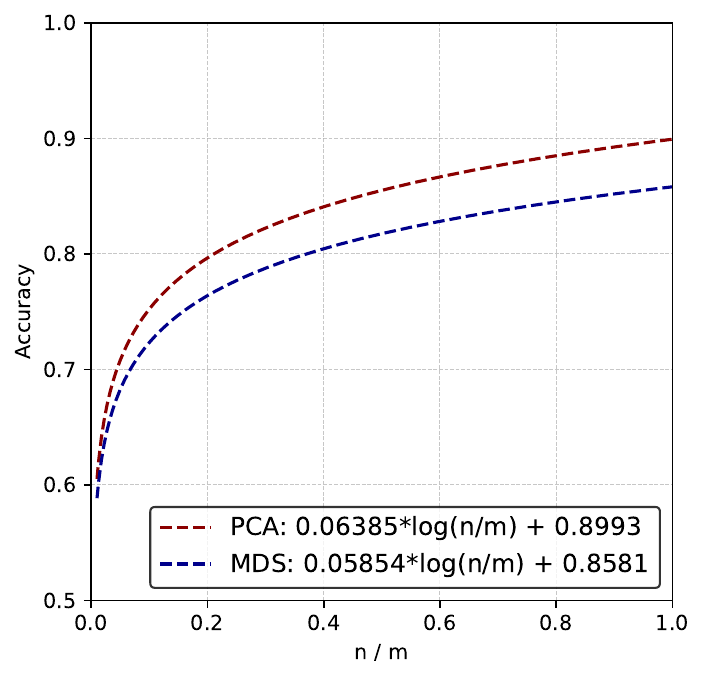}
        \caption{\small Data Fitting for Dimension Reduction on OmniCorpus dataset}\label{fig:DR_OmniCorpus}
    \end{minipage}
\end{figure}

Figure~\ref{fig:llm_fit} to   Figure~\ref{fig:OmniCorpus_model} show that using different models does not influence much the accuracy.
This implies that the closed-form function is applicable to a variety of neural network models.
Therefore, in practical applications, one should consider the nature of the data and its intended use to select the most appropriate model for experimentation.

\subsection{Influence of Dimension-Reduction Methods}
The datasets analyzed in this study exhibit extremely high dimensionality, often reaching hundreds of thousands of dimensions. Even after applying advanced embedding techniques, the resulting vectors typically retain several hundred to several thousand dimensions, presenting significant challenges for effective analysis and interpretation. To address these issues, we employed two principal dimensionality reduction methods—Multidimensional Scaling (MDS) and Principal Component Analysis (PCA)—both widely recognized for their ability to efficiently reduce data complexity while preserving the structural integrity of data relationships. As many modern dimensionality reduction techniques are fundamentally derived from PCA and MDS, our study specifically focused on these methods to assess their practical effectiveness. 

By comparing the performance of PCA and MDS on the same dataset in Figure~\ref{fig:DR}, we found that PCA is more sensitive to changes in $n/m$ and converges to higher accuracy more quickly. Additionally, in the materials science dataset, PCA achieves a peak neighborhood preservation accuracy of 100\%, surpassing the highest accuracy attainable by MDS. In the other two multimodal datasets, PCA also demonstrates greater precision compared to MDS, further validating its superiority. Moreover, while different dimensionality reduction techniques may affect specific numerical values, they do not alter the overall data pattern; the relationship between accuracy and $n/m$ remains consistent with our hypothesis.

From Figure~\ref{fig:DR} to Figure~\ref{fig:DR_OmniCorpus}, although the overall data pattern remains constant and both PCA and MDS align with our prior hypotheses, the choice of dimensionality reduction technique noticeably affects the fitting outcome. In practical applications, it is crucial to weigh the specific contexts and experimental data at hand. This assessment will inform the selection of the most appropriate dimensionality reduction method to achieve the desired level of accuracy.

\section{Conclusion and Future Work}
This paper proposes a new method, namely Order-Preserving Dimension Reduction (OPDR), to address the challenge of high-dimensional embeddings in multimodal data analytics, particularly those generated by large language models.
From a theoretical perspective, this work introduces new concepts (both point-wise and space-level) to measure the closeness of two metric spaces in terms of their $k$-nearest neighbors and a closed-form function to characterize the relationship among dimensionality, cardinality, and similarity (i.e., the accuracy of the dimension-reduction map).
The closed-form function is incorporated into a practical ecosystem by complementing other system components, such as dimension-reduction methods (PCA, MDS) and models (CLIP, ViT, Bert). 
Extensive evaluation demonstrates both the effectiveness and practical utility of OPDR.

Our future work will further explore the theoretical foundation of the closed-form function, particularly the relationship among metric spaces' dimensionality, cardinality, and the accuracy of the dimension-reduction map. The fact that the proposed order-preserving measure is a well-defined metric may provide deeper insights. Additionally, we plan to investigate the extensibility of our method in production vector database systems, such as PostgreSQL and pgvector, to support large-scale retrieval tasks. Beyond multimodal retrieval, we aim to explore potential applications in medical imaging, where high-dimensional embeddings from MRI~\cite{CHAMBERLAND201989, sridhar2022optimal}, CT scans, and vital records could benefit from dimension reduction while preserving similarity relationships—enabling efficient, accurate, and scalable content-based retrieval for medical diagnosis.



\bibliography{aaai25}

\appendix

\end{document}